\documentclass[journal]{IEEEtran}
\usepackage{amsmath}
\usepackage{booktabs}
\usepackage{amsfonts}
\usepackage{graphicx}
\usepackage{subfigure}
\usepackage{array}
\usepackage{caption}
\usepackage{cite}
\usepackage{multirow}
\usepackage{algorithm}
\usepackage{algorithmicx}
\usepackage{algpseudocode}%
\usepackage{listings}
\usepackage{amssymb}
\usepackage[normalem]{ulem}
\useunder{\uline}{\ul}{}

\begin{document}

\title{Context-aware TFL: A Universal Context-aware Contrastive Learning Framework for Temporal Forgery Localization}

\author{
	Qilin Yin,
	Wei Lu,~\IEEEmembership{Member,~IEEE,}
	Xiangyang Luo,
	Xiaochun Cao,~\IEEEmembership{Senior,~IEEE}
	
	
\thanks{This work was jointly supported by the National Natural Science Foundation of China (Grant  No. 62441237). ($Corresponding \ authors: Wei\ Lu$)}
\thanks{Qilin Yin and Wei Lu are with the School of Computer Science and Engineering, MoE Key Laboratory of Information Technology, Guangdong Province Key Laboratory of Information Security Technology, Sun Yat-sen University, Guangzhou 510006, China. (e-mail: yinqlin@mail2.sysu.edu.cn; luwei3@mail.sysu.edu.cn).}
\thanks{Xiangyang Luo is with the State Key Laboratory of Mathematical Engineering and Advanced Computing, Zhengzhou 450002, China. (e-mail: luoxy$\_$ieu@sina.com)}
\thanks{Xiaochun Cao is with the School of Cyber Science and Technology, Shenzhen Campus, Sun Yat-sen University, Shenzhen 518107, China. (e-mail:caoxiaochun@mail.sysu.edu.cn)}
}

\maketitle
\begin{abstract}
Most research efforts in the multimedia forensics domain have focused on detecting forgery audio-visual content and reached sound achievements. However, these works only consider deepfake detection as a classification task and ignore the case where partial segments of the video are tampered with. Temporal forgery localization (TFL) of small fake audio-visual clips embedded in real videos is still challenging and more in line with realistic application scenarios. To resolve this issue, we propose a universal context-aware contrastive learning framework (UniCaCLF) for TFL. Our approach leverages supervised contrastive learning to discover and identify forged instants by means of anomaly detection, allowing for the precise localization of temporal forged segments. To this end, we propose a novel context-aware perception layer that utilizes a heterogeneous activation operation and an adaptive context updater to construct a context-aware contrastive objective, which enhances the discriminability of forged instant features by contrasting them with genuine instant features in terms of their distances to the global context. An efficient context-aware contrastive coding is introduced to further push the limit of instant feature distinguishability between genuine and forged instants in a supervised sample-by-sample manner, suppressing the cross-sample influence to improve temporal forgery localization performance. Extensive experimental results over five public datasets demonstrate that our proposed UniCaCLF significantly outperforms the state-of-the-art competing algorithms.\footnote{This work has been submitted to the IEEE for possible publication. Copyright may be transferred without notice, after which this version may no longer be accessible.}

\end{abstract}

\begin{IEEEkeywords}
Multimedia forensics, temporal forgery localization, contrastive learning, anomaly detection, audio-visual consistency representation.
\end{IEEEkeywords}

\section{Introduction}\label{sec1}
Deepfake, one of the specific applications of AIGC technology, has raised profound questions about authenticity. In the face of this challenge, numerous efforts have been devoted to the multimedia deepfake detection in recent years and achieved promising performances, including audio \cite{wu2024audio,xue2023learning}, visual \cite{yin2023dynamic,zhu2024deepfake,peng2024deepfakes,zou2025semantic}, and even cross-modal forms \cite{yin2024fine,yang2023avoid}. However, in comparison to the overall forgery of multimedia content, content-driven partial multimedia forgery approaches are more cost-effective and challenging. These short modified segments can completely alter the meaning and sentiment of the original content and can evade detection by the advanced deepfake detectors mentioned above, which often assume the forged content is present throughout the visual/audio signal \cite{cai2023glitch}. Considering the widespread use and potentially pernicious effects of content-driven partial forgery of multimedia in real-world scenarios, Chugh \textit{et al.}~\cite{chugh2020not} proposed a new task called temporal forgery localization (TFL) to localize the start and end timestamps of manipulated segments. 
 
\begin{figure}[t]
	\centering
	\subfigure[A video altered by video inpainting]{
		\includegraphics[width=0.9\linewidth]{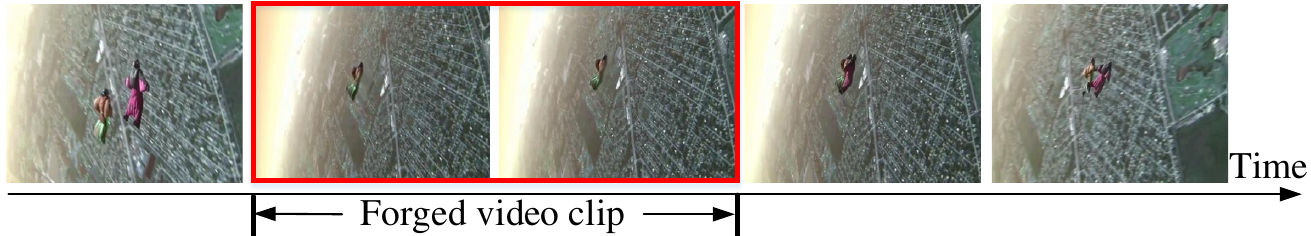}}\\
	\subfigure[Video detection results obtained by TAL methods ]{
		\includegraphics[width=0.9\linewidth]{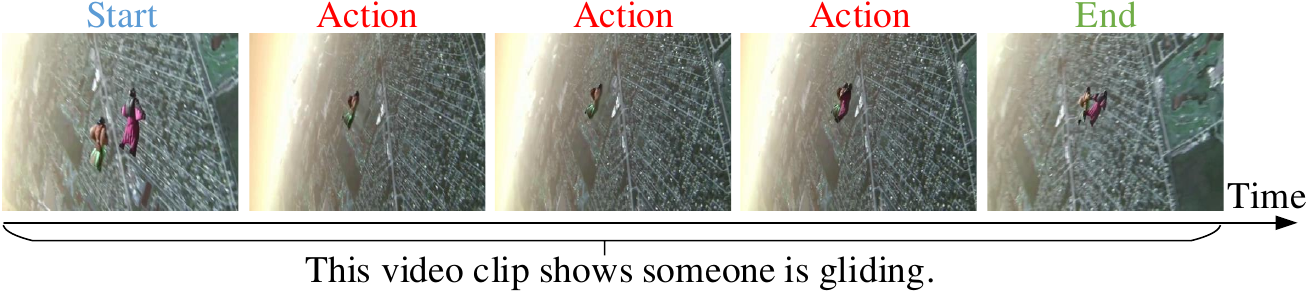}}\\
	\subfigure[Video detection results obtained by TFL methods]{
		\includegraphics[width=0.9\linewidth]{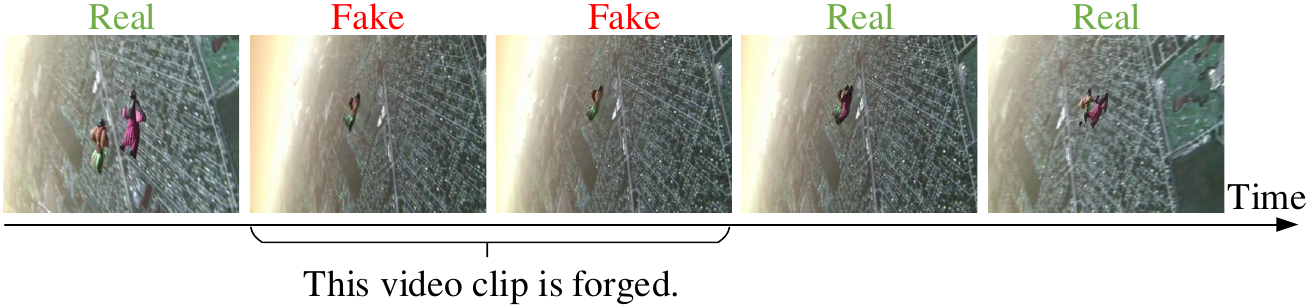}}
	\caption{Examples of video detection results for different types of methods. Regardless of tampering, the TAL method identifies that the entire video depicts someone gliding, while the TFL method can provide the temporal interval of the tampered clip of the video.}
	\label{fig:example}
\end{figure} 

Although TFL is similar to temporal action localization (TAL) in that both aim to give the temporal interval of a specific instance and its corresponding label, the two tasks are inherently different. TAL methods primarily focus on semantic perception of the video, detecting and classifying specific actions within the video. TFL methods are designed to capture abnormal information in the video that indicates tampering. TAL involves learning and perceiving high-level semantic information in the video, while TFL focuses on modeling and analyzing low-level abnormal information in the video. As shown in Fig.~\ref{fig:example}, TAL method classify the entire video into `gliding' class, regardless of whether the video content is real or AI-generated. In contrast, TFL methods can differentiate between real and fake segments within the video, identifying the exact intervals where forgery has occurred.  

TFL tasks present unique challenges compared to TAL. Therefore, TAL methods may not be suitable for TFL tasks. Recently, a few studies have been devoted to TFL task. Both BA-TFD+ \cite{cai2023glitch} and AVTFD \cite{liu2023audio} were designed for multimodal forgery localization, ignoring the scenario of unimodal (audio or video) forgery localization. Although they combined frame-level deepfake detection approaches with an anchor-free temporal regression header, achieving some degree of localization performance, their modeling of abnormal information was insufficient for obtaining discriminative instant features. DiMoDif \cite{koutlis2024dimodif} improves localization accuracy by using feature pyramids and local attention mechanisms. Both UMMAFormer \cite{zhang2023ummaformer} and MFMS \cite{zhang2024mfms} captured abnormal information through feature reconstruction. Moreover, MFMS adopted cross-attention to fuse different modalities and features. However, in content-driven partial forgery scenarios, the tampered region usually accounts for only a small fraction of the forgery samples in either the temporal or spatial dimensions, which leads to the problem of local anomalous information sparsity. The video-based feature reconstruction approach is not sensitive to such tiny anomalous information. Therefore, the distinguishability of the instant features extracted by the aforementioned methods is not sufficient to achieve more precise temporal forgery localization.

To overcome the shortcomings in the existing studies, we propose a universal context-aware contrastive learning framework (UniCaCLF) for extracting anomalous information from input multimedia and generating discriminative instant features for TFL tasks. The underlying assumption is that in content-driven partial forgery scenarios, forged instant features tend to be more distant from the majority of instant features (\textit{i.e.}, global context) than genuine instant features. Therefore, we attempt to compare each instant feature with the global context of the input multimedia and identify abnormal instants (\textit{i.e.}, forged instants) based on their distances. Notably, we construct the context-aware contrastive objective in a supervised sample-by-sample manner. This means that the positive-negative contrast relationship between genuine and forged instants is confined within the same sample, effectively avoiding the interference from inconsistent distribution of anomalous information caused by different tampering operations across samples. The proposed framework focuses on capture the abnormal feature in a given input feature and is suitable for arbitrary modal inputs.

Specifically, we propose a novel context-aware perception layer (CaP layer) consisting of the heterogeneous activation operation (HAO) and an adaptive context updater (ACU). HAO can selectively enhance the representation of forged instants based on the similarity of each instant feature with respect to the global context, thus effectively improving the distinguishability between instant features and mitigating the problem of rank loss \cite{dong2021attention} due to the self-attention. HAO is developed around the global context, and a precise and positive global context facilitates improved performance in detecting forged instants. Therefore, ACU is designed to mitigate the negative influence of forged instants in the global context generation process as much as possible through a memory-based context update mechanism. Furthermore, a context-aware contrastive loss (CaCL) is designed in a supervised sample-by-sample manner. This design allows the proposed UniCaCLF to be immune from the mutual influence of features across different samples in a batch, and focuses on capturing abnormal information within each sample, which effectively amplifies the discrepancy between fake and genuine instants. It achieves this by maximizing the consistency between positive pairs (genuine instant features and global context) and minimizing the consistency between negative pairs (forged instant features and global context). 

Extensive experiments demonstrate the effectiveness and generalizability of UniCaCLF for the different TFL tasks on five benchmark datasets, Lav-DF \cite{cai2023glitch}, AV-Deepfake1M\cite{cai2024av}, TVIL \cite{zhang2023ummaformer}, HAD \cite{yi2021half} and Psynd \cite{zhang2022localizing}.  Our contributions can be summarized as follows:
\begin{itemize}
	\item We propose an universal context-aware contrastive learning framework for temporal forgery localization. This framework concentrates on modeling the distribution of anomalous features within a sample, achieving efficient temporal forgery localization through the context-aware contrastive coding.
	\item A context-aware perception layer is proposed to generate discriminative instant features and the positive global context by means of anomaly detection.
	\item A context-aware contrastive loss is devised to mine intra-sample abnormal information in a supervised sample-by-sample manner, suppressing the cross-sample influence and further extending the discrepancy between fake and genuine instants.
\end{itemize}
The rest of this paper is organized as follows: Section \ref{sec:related} introduces some relative domain knowledge, Section \ref{sec:method} elaborates the proposed UniCaCLF, Section \ref{sec:exper} shows experiments, and Section \ref{sec:con} concludes the paper.

\begin{figure*}[htbp]
	\begin{center}
		\includegraphics[width=0.8\linewidth]{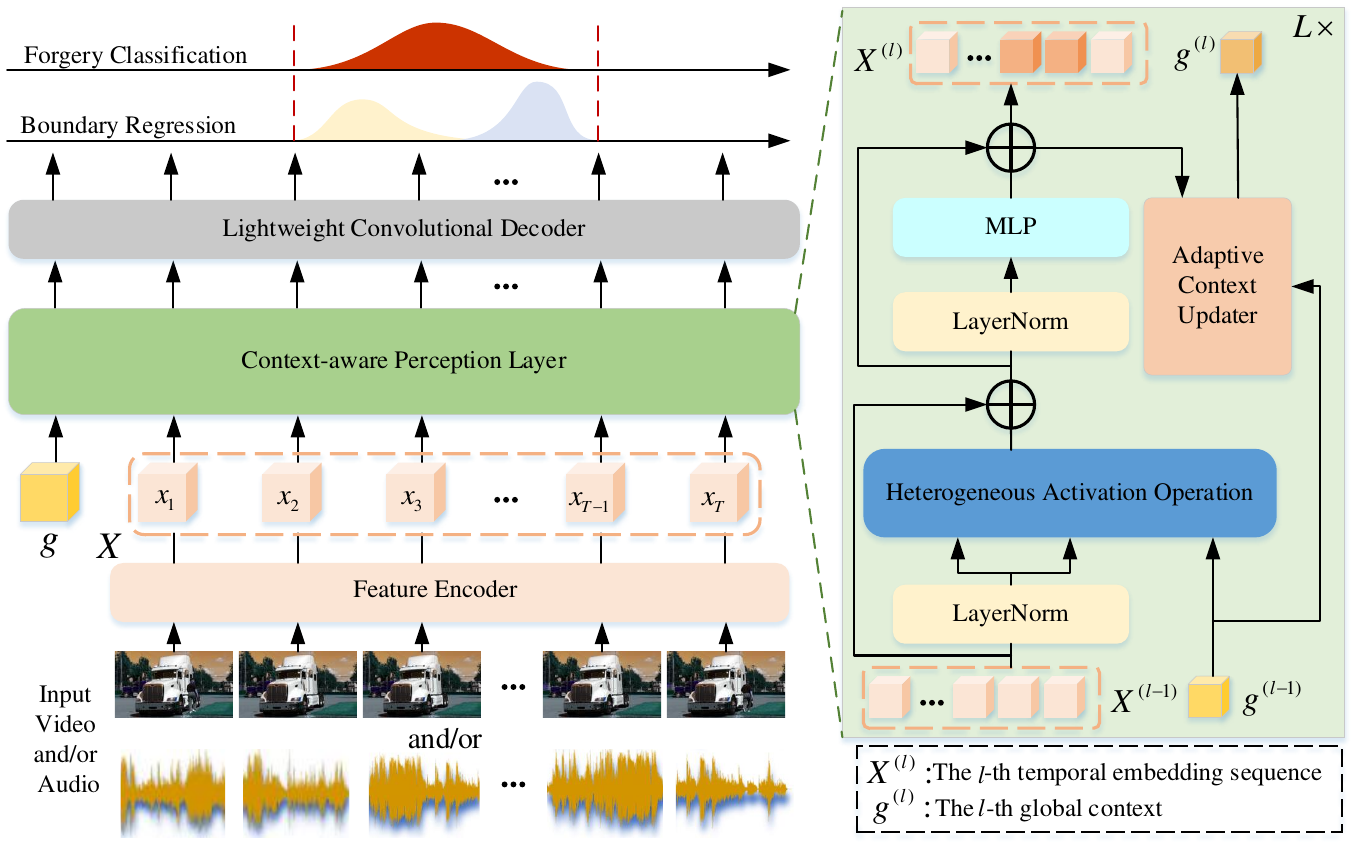}
	\end{center}
	\caption{Diagrammatic overview of the proposed universal context-aware contrastive learning framework.}
	\label{fig:model}
\end{figure*}
 
\section{Related Work}
\label{sec:related}
\subsection{Multi-modal Deepfake Detection}
With more research being done in the field of Deepfake detection, numerous approaches are now concentrating on the multimodal approach by utilizing both visual and audio information. Most of them \cite{mittal2020emotions, hashmi2022multimodal, she2024using, zhou2021joint, ilyas2023avfakenet, yin2024fine, yang2023avoid} identify Deepfakes by intrinsically measuring the degree of audio-visual correspondence. For example, EmoForen \cite{mittal2020emotions} proposed detecting Deepfake videos by analyzing audio-visual cues and perceiving emotion clues. Hshmi \textit{et al}. \cite{hashmi2022multimodal} fed both visual features and audio feature into multiple learners, producing the final prediction result by hard voting. Nevertheless, these methods are limited to determining whether the audio and visual components are consistent, without specifically identifying which modality has been manipulated. JointAV \cite{zhou2021joint} and AVFakeNet \cite{ilyas2023avfakenet} combined the concurrency property of audio and visual feature with a multi-classifier approach to detect manipulation in each modality. Yin \textit{et al}. \cite{yin2024fine} leveraged a heterogeneous graph to model intra- and inter-modality relationships, achieving fine-grained classification results. These approaches typically require training the model with ground truth labels of fake and real videos. In contrast, POI-Forensics \cite{cozzolino2023audio} designed the ID-identity feature encoder, which was trained by the sequences of synchronization characteristics taken from only real videos to extract discriminative features.

\subsection{Temporal Action Localization}
Temporal action localization (TAL) is a critical task for video understanding that aims to detect the boundaries and categories of action segments in an untrimmed video \cite{zhu2023contextloc++}. Existing temporal action detection methods can be divided into two categories: anchor-based methods and anchor-free methods \cite{vahdani2022deep}.

Anchor-based methods generate proposal based on a pre-determined set anchors. Existing anchor-based works can be categorized as the one-stage pipeline \cite{fu2023semantic, long2019gaussian, rahman2020mid} and two-stage pipeline \cite{xu2017r, chao2018rethinking, li2020graph}. The one-stage pipeline predicts temporal boundaries and action categories simultaneously but is limited by anchor designs and requires fine-tuning. In contrast, the two-stage pipeline first generates coarse proposals, then refines them, offering better performance but with more complex designs.

In contrast to anchor-based methods that rely on default anchors, anchor-free methods directly learn to predict temporal proposals without referencing default anchors. For example, BSN \cite{lin2018bsn} predicted the start, end and action category at each temporal location and generated proposals using locations with high start and end probabilities. Afterwards, BMN \cite{lin2019bmn} additionally generated a boundary-matching confidence map to improve proposal generation. More recently, ActionFormer \cite{zhang2022actionformer} pushes the localization performance to a brand-new status and TriDet \cite{shi2023tridet} has developed it even further. While no pre-defined anchor boxes are required, these methods often encounter challenges related to the center uncertainty problem \cite{wang2023temporal}.

Both TAL and TFL follow a similar detection process: (1) projecting the original inputs to the feature sequence; (2) constructing multi-scale instant representations to facilitate the alignment between the video features and downstream task; (3) decoding the representations with the detection head to generate the temporal interval of the specific instance and its corresponding label. However, TFL requires more granular instant representations than TAL. TAL aims to identify semantic differences visible to the naked eye, such as different categories of actions. In contrast, TFL is more focused on detecting subtle and often invisible forgery traces within the same category of content.

\subsection{Contrastive Learning}
Contrastive learning is a feature representation learning method that enhances the generalization and robustness of features by comparing similarities and differences between different samples, thus improving the performance of the model. This methodology has been recently popularized for un-/self-supervised representation learning \cite{chen2020simple, chen2021exploring, he2020momentum}. SimCLR \cite{chen2020simple} proposed a simple contrastive learning framework by augmenting and embedding the same image to obtain two versions of feature representations in the contrasting space, which are subsequently combined with NT-Xent \cite{oord2018representation} to obtain robust visual representations. In practice, contrastive learning methods benefit from a large number of negative samples \cite{wu2018unsupervised}. Given that SimCLR's performance is limited by the size of the batch, Moco \cite{he2020momentum} proposed a memory bank to maintain a large of negative samples, unbinding the constraint of batch size.

UniCaCLF that performs contrastive learning in a supervised learning manner, differs from existing contrastive learning frameworks \cite{qiao2024fully, li2021frequency, sun2022dual} for Deepfake detection, because most of them are self-supervised and the contrastive instances should be constructed from the unlabeled data. For example, FUDD \cite{qiao2024fully} and DCL \cite{sun2022dual} contrast between different data augmentations. Given the embedding of a face as the query, the positive key is the embedding of the same face after different data augmentations, while the negative keys are those from different faces. In contrast, in our UniCaCLF, given the global context of the input as the query, the positive and the negative keys are constrained to the genuine and forged instants of the same input, respectively. In light of this, UniCaCLF is a kind of supervised contrastive learning manner \cite{khosla2020supervised}.

\section{Universal Context-aware Contrastive Learning Framework}
\label{sec:method}
\subsection{Problem Definition}
We first give a formal definition of the TFL task. Specifically, given an untrimmed temporal data sequence
$\mathcal{D}$, we have a set of temporal embedding feature ${X} = \{ {x_t}\} _{t = 1}^{{T}}$ from the temporal sequence $\mathcal{D}$, where $T$ corresponds to the number of instants, and $K$ segments $Y = (s_k, e_k),k=1,2,...,K$ with the forged segment start instant $s_k$ and the end instant $e_k$. TFL aims to locate all segments $Y$ based on the input feature $X$. $\mathcal{D}$ can be extended to visual-only, audio-only, audio-visual modalities or other types of multimedia sequence. 

\subsection{Overview}
In order to establish a straightforward and universal framework to facilitate the analysis and development of TFL tasks, we propose a universal context-aware contrastive learning framework (UniCaCLF), as shown in Fig.~\ref{fig:model}. UniCaCLF aims to identify an instant's abnormality by comparing it to the global context (\textit{e.g.}, the average of all instants). It is worth noting that the global context will be re-estimated later, as detailed in Section~\ref{sec:CaP}. The core concept is that an instant that strays significantly from the majority in the feature space is likely to be a forgery in context-driven partial forgery scenarios. 

Specifically, given an arbitrary training data $\mathcal{D}$ with $T$ instants, a pre-trained feature extractor (\textit{e.g.}, TSN \cite{wang2016temporal} or I3D \cite{carreira2017quo} for video modality, or BYOL-A \cite{niizumi2021byol} for audio modality) is first used to extract the corresponding instant-level features ${X} = \{ {x_t}\} _{t = 1}^{{T}}$. Following that, a CaP feature pyramid is built to tackle forged segments with various temporal lengths. In other words, the temporal instant feature $X$ is iteratively downsampled and processed $L$ times with the context-aware perception (CaP) layer to achieve a discriminative feature set $\left\{ {{X^{(l)}}} \right\}_{l = 1}^L$ consisting of different scale level temporal instant features and a corresponding global context sets $\left\{ {{g^{(l)}}} \right\}_{l = 1}^L$. Finally, the CaP feature pyramid is shared by the classification and regression heads to accomplish the detection task and generate the temporal interval of the forged segment with its corresponding label.

It is worth noting that we also design a context-aware contrastive loss (CaCL) to constrain the UniCaCLF to focus on intra-sample abnormal information induced by tampering operations, which further improves the distinguishability of the temporal instant feature sequences. In the below we introduce CaP layer and CaCL in Sections \ref{sec:CaP} and \ref{sec:CaCL}, respectively.

\begin{figure*}[t]
	\begin{center}
		\includegraphics[width=0.8\linewidth]{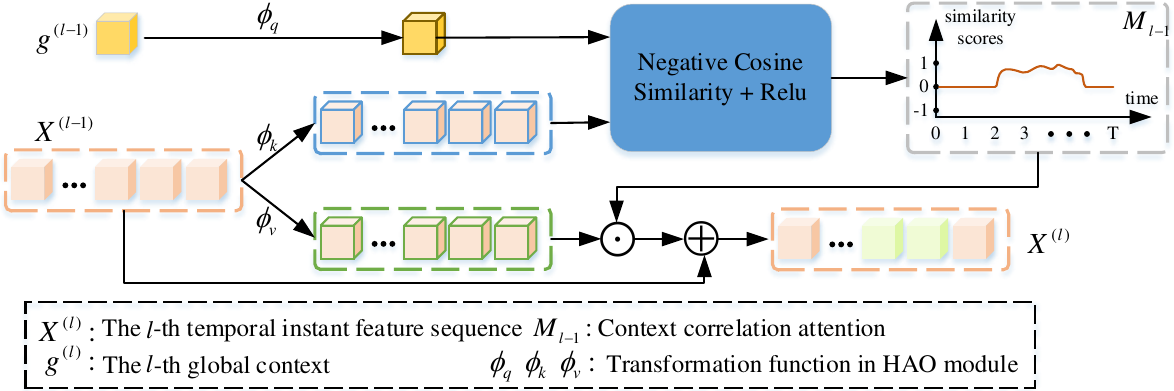}
	\end{center}
	\caption{Illustration of the structure of HAO. $\oplus$ and $\odot $ denote the element-wise addition and element-wise multiplication respectively.}
	\label{fig:cap}
\end{figure*}

\subsection{Context-aware Perception Layer}
\label{sec:CaP}
Given an arbitrary untrimmed sample $\mathcal{D}$, the key task of UniCaCLF is to extract discriminative instant features from the temporal instant feature sequences $X$ for TFL. To achieve this goal, our proposed UniCaCLF should be sensitive to tiny abnormal information induced by tampering operations and have the ability to discover and identify forged instants, which can facilitate the localization of those forged segments in $\mathcal{D}$. Consequently, we construct the CaP layer to identify forged instant features from $X$ by means of anomaly detection and enhance the distinguishability between instant features. Specifically, given the temporal instant feature sequence $X^{(l-1)}$ and global context $g^{(l-1)}$, the heterogeneous activation operation (HAO) first builds the contrastive objective based on global context to find and enhance forged instant features according to their distances from the global context. Then, the updated temporal instant feature sequences $X^{(l)}$ are fed into the adaptive context updater (ACU) to generate a more positive global context $g^{(l)}$ that is not negatively affected by the forged instant features. HAO and ACU cooperate with each other can effectively enhance the distinguishability between genuine and forged instant features. The following content details the specific operations.

\textbf{Heterogeneous activation operation}. As shown in Fig.~\ref{fig:cap}, HAO first utilizes three transformation functions (${\phi _q}( \cdot )$, ${\phi _k}( \cdot )$ and ${\phi _v}( \cdot )$) to embed $g^{(l-1)}$ and $X^{(l-1)}$ into query, key and value feature, respectively. Then, the HAO calculates the negative cosine similarity of each instant feature in ${\phi _k}( X^{(l-1)} )$ to ${\phi _q}( g^{(l-1)} )$ and activates it to measure the context correlation between each instant feature and the global context. A greater activation value means that the corresponding instant is most likely a forged instant. Depending on the context correlation attention $M_{l-1}$ obtained by collecting all activation values in order, HAO can effectively find and strengthen forged instant features, thus achieving the updated temporal instant feature sequence $X^{(l)}$. Mathematically, the HAO can be written as:
\begin{normalsize}
	\begin{equation}
	\label{eqn:01}
	\begin{aligned}
{X^{(l)}} & = {f_{HAO}}({g^{(l - 1)}},{X^{(l - 1)}}) \\
& = {\mathop{\rm Re}\nolimits} {\rm{LU}}({\rm{Sim}}({\phi _q}({g^{(l - 1)}}),{\phi _k}({X^{(l - 1)}})) \times {\phi _v}({X^{(l - 1)}}) \\
& \quad + {X^{(l - 1)}}
\end{aligned}
	\end{equation}
\end{normalsize}
where $\rm{Sim}(\cdot)$ is the calculation of negative cosine similarity. It is worth noting that HAO uses $\rm{Relu}$ as activation function, which means that if the activation value is greater than 0, the corresponding instant feature (\textit{i.e.}, forged instant features) is augmented, while the rest of the instant features remain unchanged.

\textbf{Adaptive context updater}. Throughout the above description, the global context plays a critical role in the UniCaCLF, which is an important part of the context-aware contrastive objective. The precision of the global context has a great impact on the identify accuracy of the forged instants and improving the discriminability of instant feature.
\begin{figure}[t]
	\begin{center}
		\includegraphics[width=0.9\linewidth]{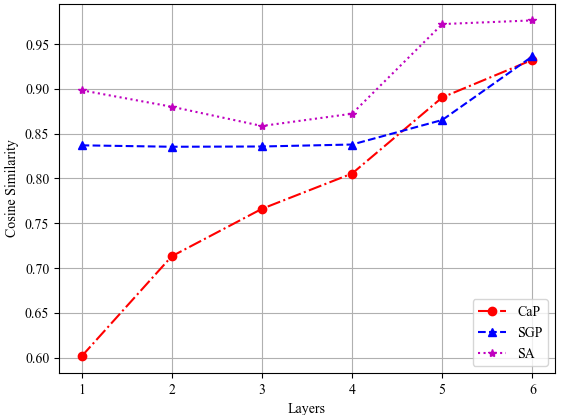}
	\end{center}
	\caption{Within the TVIL dataset and TSN pre-trained features, we statistic the average cosine similarity between each instant feature and the corresponding video-level average feature (\textit{i.e.}, the average of all instant features) across each level of the feature pyramid generated by self-attention (SA), SGP layer and CaP layer, respectively. It is observed that CaP layer exhibits low similarity, which means CaP can extract more discriminative instant features.} 
	\label{fig:cos}
\end{figure}
Straightforwardly, we can perform sum, max, or average pooling of the temporal instant feature sequence $X^{(l)}$ to obtain the global context. However, these operations do not distinguish forged and genuine instants, which results in the inaccurate global context. To alleviate it, ACU introduces a memory buffer to register the current global context (\textit{i.e.}, $g^{(l-1)}$), based on which ACU calculates the contribution of each instant feature in $X^{(l)}$ to $g^{(l-1)}$.  The assumption is an instant feature that is deviated from the current global context should be weakened in the next CaP layer. Formally,
\begin{normalsize}
	\begin{equation}
		\label{eqn:03}
		s_t^{(l)} = {\rm{cosine}}(x_t^{(l)},{g^{(l - 1)}}),
	\end{equation}
\end{normalsize}
\begin{normalsize}
	\begin{equation}
		\label{eqn:04}
		\alpha _t^{(l)} = \frac {\exp (s_t^{(l)})} {\sum\nolimits_{t = 1}^T {\exp (s_t^{(l)})} },
	\end{equation}
\end{normalsize}
\begin{normalsize}
	\begin{equation}
		\label{eqn:05}
		{g^{(l)}} = \beta \cdot {g^{(l-1)}} + (1-\beta) \cdot  \sum\limits_{t = 1}^T {\alpha _t^{(l)} \cdot x_t^{(l)}} 
	\end{equation}
\end{normalsize}
where $\beta$ is a trainable parameter and $T$ denotes the number of the instant features. First, ACU computes the attention $\alpha _t^{(l)}$ for each instant feature based its cosine similarity with the current global context $g^{(l-1)}$. Then, ACU generates the new global context $g^{(l)}$ by weighted aggregated all instant features and $g^{(l-1)}$. Following this approach, ACU can reduce the negative influence of noisy instant feature and generate the more positive global context.

Recently, UMMAFormer \cite{zhang2023ummaformer} adopts the transformer-based feature for TFL and shows promising performance. However, TriDet \cite{shi2023tridet} demonstrated that self-attention (SA) mechanism of transformer layer leads to the rank loss problem across the temporal dimension, \textit{i.e.}, instant features extracted via SA tend to exhibit high similarities between instants. To mitigate the problem, TriDet proposed a Scalable-Granularity Perception (SGP) layer that injects instant features with different temporal scope semantic information, thus increasing the discriminability of each instant feature. However, this is not applicable to TFL task, because in context-driven partial forgery scenarios, the macro-semantic information in the forgery samples always remains constant, which is not conducive to SGP layer. In contrast, CaP layer is excellent at capturing tiny anomalous information in forgery samples and can provide discriminative instant features. To verify the validity of the CaP layer, we statistic the average cosine similarity of instant features extracted by SA, SGP and CaP, respectively. As shown in Fig.~\ref{fig:cos}, we can observe that the CaP layer indeed decreases the similarity between each instant feature and the video-level average feature compared to SA and SGP. In other words, CaP layer can effectively increase the distinguishability between instant features. In addition, CaP is much more efficient compared to the dense calculation of SA and SGP.

\subsection{Context-aware Contrastive Loss}
\label{sec:CaCL}
Although the CaP layer can increase the distinguishability of instant features, it only enlarges the distance from the forged instants to the global context and ignores genuine ones. In order to further enlarge the feature dissimilarity between forged and genuine instant, we can compress the distance from the genuine instants to the global context. Therefore, we introduce the context-aware contrastive loss (CaCL) to supervise UniCaCLF to learn and extend the discrepancy between genuine and forged instants in a sample-by-sample contrastive learning manner. The goal of CaCL is to narrow the distance between the genuine instant feature and the global context, while enlarge the distance between the forged instant feature and the global context. 

Specifically, given the global context $g$ and the temporal instant feature sequences $X = \{ {x_t}\} _{t = 1}^T$ and the corresponding instant-level labels $Y=\{ {y_t}\} _{t = 1}^T, {y_t} \in \{ 0,1\} ,\forall t$ where ${y_t}=1$ indicates the corresponding instant is forged, CaCL first divides $X$ into two feature subsets $\{ x_1^ + ,x_2^ + ,x_3^ + , \cdot  \cdot  \cdot ,x_J^ + \} $ and $\{ x_1^ - ,x_2^ - ,x_3^ - , \cdot  \cdot  \cdot ,x_K^ - \} $. $\{ x_1^ + ,x_2^ + ,x_3^ + , \cdot  \cdot  \cdot ,x_J^ + \} $ represents the instant features belonging to genuine segments, while $\{ x_1^ - ,x_2^ - ,x_3^ - , \cdot  \cdot  \cdot ,x_K^ - \} $ stands for those of forged segments. Then, CaCL treats $g$ as the query and compute the intra-sample contrastive loss as follows:
\begin{normalsize}
	\begin{equation} 
	\label{eqn:02}
	\begin{aligned}
		&{Loss_{intraCL}} \\
		&=  - \log {{{1 \over J}\sum\nolimits_{j \in [1,J]} {\exp (g \cdot x_j^ + /\tau )} } \over {{1 \over J}\sum\nolimits_{j \in [1,J]} {\exp (g \cdot x_j^ + /\tau )}  + \sum\nolimits_{k \in [1,K]} {\exp (g \cdot x_k^ - /\tau )} }}
	\end{aligned}
	\end{equation}
\end{normalsize}
where $\tau$ is a temperature hyper-parameter. As Eq.~(\ref{eqn:02}) shown, CaCL calculates the mean distance from all genuine instant feature to the global context and attempts to compress it, which facilitates the optimization process. It should be emphasized that CaCL is calculated in a sample-by-sample manner (one-by-one) for each sample in the forward mini-batch, rather than over the entire batch. This is because the feature discrepancy between genuine and forged instants in each sample varies depending on the types of manipulation method. Existing forgery datasets all contain multiple tampering methods, with each sample undergoing a different kind of tampering. Learning cross-sample feature discrepancies tends to caused UniCaCLF confusion and is not conducive to mine general anomaly features. Thus, given a mini-batch sample containing $B$ samples, the overall context-aware contrastive loss ($Loss_{CaCL}$) is calculated by
\begin{normalsize}
	\begin{equation}
		\label{eqn:06}
		{Loss_{CaCL}} = {1 \over B}\sum\limits_{i = 1}^B {{Loss_{{\mathop{\rm int}} raCL}}({g_i},{X_i})} 
	\end{equation}
\end{normalsize}
With the supervision of $Loss_{CaCL}$, UniCaCLF can learn more discriminative instant features and have generalization capability.

\begin{algorithm}[t]
	\caption{UniCaCLF}\label{algorithm}
	\begin{algorithmic}[1]
		\Require A set of labeled temporal embedding feature ${X} = \{ {x_t}\} _{t = 1}^{{T}},{x_t} \in {\mathbb{R}^{C}},\forall t$, the corresponding instant-level labels $Y=\{ {y_t}\} _{t = 1}^T, {y_t} \in \{ 0,1\} ,\forall t$ and the learning rate $\eta$.
		\Ensure Learned parameter $W$ of UniCaCLF.
		\State Initialize the temporal instant feature sequence ${X^{(0)}} = X$
		\State Initialize the global context ${g^{(0)}} = {1 \over T}({\sum\nolimits_{t = 1}^T x _t})$
		\Repeat
		\For{$l=1:L$}
		\State ${X^l} = {f_{HAO}}({g^{(l - 1)}},{X^{(l - 1)}})$ 
		\State $g^{l} = {f_{ACU}}({g^{(l - 1)}},{X^{(l)}})$
		\EndFor
		\State Compute the loss function in Eq. \ref{eqn:07}
		\State Update $W$ by $W= W - \eta {\nabla _W}L$
		\Until {Converges}
	\end{algorithmic}
\end{algorithm}

\subsection{Training and Inference}
Each scale level of the feature pyramid outputs a temporal instant feature sequence, which is then fed into the classification head and regression head to localize forged segments. The overall loss function is defined as follows:
\begin{normalsize}
	\begin{equation}
	\label{eqn:07}
	Loss = {Loss_{cls}} + \varphi_{1}{Loss_{reg}} + \varphi_{2} {Loss_{CaCL}},
	\end{equation}
\end{normalsize}
where ${Loss_{cls}}$ is the classification loss and $Loss_{reg}$ is the regression loss. $\varphi_{1}$ and $\varphi_{2}$ are hyperparameters to balance the relationship between different losses. By default, we set $\varphi_{1} = 2$ and $\varphi_{2} = 0.5$. Since the forged segment usually occupies a small portion of an untrimmed video, we use the focal loss \cite{lin2017focal} as the classification loss because it can solve the problem of fake and genuine sample imbalance. Meanwhile, Distance-IoU (DIoU) loss \cite{zheng2020distance} is adopted as the regression loss for faster convergence during training and more accurate boundary. Algorithm \ref{algorithm} outlines the training process. 

In inference phase, our model output $({p_i},{s_i},{e_i})_{i = 1}^n$ as the prediction forgery set, where $n$ denotes the total number of prediction forged segments. $s_i$, $e_i$ represent the start time and end time of the $i$-th forged segment and $p_i$ is its classification confidence. Soft-NMS \cite{bodla2017soft} is applied to remove the redundant forged segments.

\begin{table*}[t]
	\caption{Temporal forgery localization results on the LAV-DF dataset. Best results are in bold and second best underlined.}
	\centering
	\label{tab:fullset}
	\resizebox{0.9\textwidth}{!}{
		\fontsize{6}{8}\selectfont
		\begin{tabular}{c|cccc|ccccc}
			\hline
			\multirow{2}{*}{Method} & \multicolumn{4}{c|}{AP@IoU(\%)}                   & \multicolumn{5}{c}{AR@Proposals(\%)} \\
                        & 0.5 & 0.75 & 0.95 & \multicolumn{1}{c|}{Avg.} & 100   & 50   & 20   & 10   & Avg.   \\ \hline
			MDS \cite{chugh2020not}         & 12.78          & 1.62           & 0.00           & 4.80            & 37.88          & 36.71          & 34.39          & 32.15          & 35.28          \\
			AVFusion \cite{bagchi2021hear}     & 65.38          & 23.89          & 0.11           & 29.79          & 62.98          & 59.26          & 54.80           & 52.11          & 57.29          \\
			AVTFD \cite{liu2023audio} & 94.89          & 74.87          & 1.94           & 57.23          & 76.12          & 74.24         & 72.17           & 71.24          & 73.44          \\
			BA-TFD+ \cite{cai2023glitch}     & 96.30           & 84.96          & 4.44           & 61.90           & 81.62          & 80.48          & 79.40           & 78.75          & 80.06          \\
			ActionFormer \cite{zhang2022actionformer} & 97.10           & 89.52          & 33.6           & 73.41          & {\ul92.59}          & {\ul 92.03}          & 91.26          & 90.21          & {\ul 91.52}         \\
			TriDet \cite{shi2023tridet}         & 96.29          & 86.84          & 23.64          & 68.92          & 91.00          & 90.39          & 89.71          & 88.69          & 89.95          \\
			UMMAFormer \cite{zhang2023ummaformer}   & {\ul 98.34}          & {\ul 93.54}          & {\ul 37.22}          & {\ul 76.37}          & 91.67          & 91.60           & {\ul 91.34}          & {\ul 90.79}          & 91.35          \\ 
			MFMS \cite{zhang2024mfms} 	&\textbf{98.47}  &\textbf{94.15}  &27.80   &73.47   &90.69  &90.65  &90.46 &90.02  &90.46 \\
			\hline
			UniCaCLF       & 97.81       & 93.11          & \textbf{53.61}          & \textbf{81.51}          & \textbf{94.58}          & \textbf{94.02}          & \textbf{93.82}          & \textbf{93.16}          & \textbf{93.80}         \\ \hline
		\end{tabular}
	}
\end{table*}

\begin{table*}[t]
	\caption{Temporal forgery localization results on the AV-Deepfake1M dataset. Best results are in bold and second best underlined.}
	\centering
	\label{tab:1m}
	\resizebox{0.9\textwidth}{!}{
		\fontsize{8}{10}\selectfont
		\begin{tabular}{c|cccc|cccccc}
			\hline
			\multirow{2}{*}{Method}  & \multicolumn{4}{c|}{AP@IoU(\%)}                         & \multicolumn{6}{c}{AR@Proposals(\%)} \\         
			   & 0.5 & 0.75 & 0.95 & \multicolumn{1}{c|}{Avg.} & 50   & 30   & 20   & 10 &5  & Avg.   \\ \hline
			BA-TFD+ \cite{cai2023glitch} &74.88&59.35&4.36&46.19&75.59&73.53&71.71&68.04&6364&70.50         \\
			ActionFormer \cite{zhang2022actionformer} &97.08&91.13&36.76&74.99&92.28&91.95&91.53&90.23&88.12&90.82     \\
			TriDet \cite{shi2023tridet}  &88.07&74.92&10.28&57.76&82.24&81.03&80.10&77.85&74.75&79.19                     \\
			UMMAFormer \cite{zhang2023ummaformer} &{\ul98.30}&{\ul94.30}&{\ul41.17}&{\ul 77.92}&{\ul 92.85}&{\ul 92.52}&{\ul 92.15}&{\ul 91.21}&{\ul89.72}&{\ul91.69}              \\ 
			MFMS \cite{zhang2024mfms} &95.62&87.82&29.32&70.92&89.84&89.16&88.45&86.77&84.95& 87.83	 \\
			\hline
			UniCaCLF  &\textbf{99.03}&\textbf{94.80}&\textbf{52.71}&\textbf{82.18}&\textbf{94.92}&\textbf{94.80}&\textbf{94.64}&\textbf{93.92}&\textbf{92.32}&\textbf{94.12}                \\ \hline
		\end{tabular}
	}
\end{table*}

\begin{table*}[t]
	\caption{Temporal forgery localization results on the TVIL dataset. Best results are in bold and second best underlined.}
	\centering
	\label{tab:tvil}
	\resizebox{0.9\textwidth}{!}{
		\fontsize{6}{8}\selectfont
		\begin{tabular}{c|cccc|ccccc}
			\hline
			\multirow{2}{*}{Method}  & \multicolumn{4}{c|}{AP@IoU(\%)}                                    & \multicolumn{5}{c}{AR@Proposals(\%)}                                                 \\
			          & 0.5            & 0.75           & 0.95           & Avg.           & 100            & 50             & 20             & 10             & Avg.            \\ \hline
			ActionFormer \cite{zhang2022actionformer}                               & 84.58          & 81.81          & {\ul 70.91}    & 79.10          & {\ul 92.82}    & {\ul 90.84}    & 88.32          & 87.11          & {\ul 89.77}   \\
			TriDet \cite{shi2023tridet}                                   & 84.40          & 79.80          & 62.74          & 75.65          & 91.12          & 88.84          & 86.03          & 83.10          & 87.27         \\
			UMMAFormer \cite{zhang2023ummaformer}                                & \textbf{89.40} & \textbf{85.99} & 62.66          & {\ul 79.35}    & 91.08          & 90.58          & {\ul 88.51}    & {\ul 87.16}    & 89.33         \\ \hline
			UniCaCLF                                   & {\ul 87.07}    & {\ul 84.75}    & \textbf{74.99} & \textbf{82.27} & \textbf{92.93} & \textbf{91.83} & \textbf{90.15} & \textbf{88.43} & \textbf{90.84} \\ \hline
		\end{tabular}
	}
\end{table*}

\section{Experiments}\label{sec:exper}
\subsection{Experimental Setup}
\subsubsection{Datasets}
We conduct experiments on four challenging datasets for three forgery scenarios, including visual forgery, audio forgery, and audio-visual forgery.
\begin{itemize}
	\item LAV-DF \cite{cai2023glitch} is a large-scale multimodal forgery dataset, consisting of 36431 real videos and 99873 fake videos. The fake videos are partially forged and the duration of fake segment is in the range of $[0-1.6]$ seconds. There are four types data (full truth, visual-only tampering, audio-only tampering, and audio-visual tampering) in LAV-DF. 
	\item AV-Deepfake1M \cite{cai2024av} is a large-scale multimodal forgery dataset which
	contains 2,068 subjects resulting in 286K genuine videos and 860K forged videos. It is worth that the metadata for its test samples has not been released. Therefore, we train the models on the training set and test them on validation set.
	\item TVIL \cite{zhang2023ummaformer} is a visual-only modified dataset containing 914 real videos and 3539 partially fake videos. Compared to LAV-DF, TVIL focuses on more generalized scenarios, not just on faces.
	\item HAD \cite{yi2021half} is an audio-only modified dataset consists of training 26554 utterances, 8914 development utterances and 9072 test utterances. The fake audio in the HAD dataset involves only changing a few words in an utterance.
	\item Psynd \cite{zhang2022localizing} is a multi-speaker English corpus of approximately 13 hours in total at 24kHz sampling rate read English speech injected with synthetic speech. 
\end{itemize}

\subsubsection{Evaluation Metrics}
For evaluation, we use Average Precision (AP) and Average Recall (AR) as the evaluation metrics following \cite{cai2023glitch, zhang2023ummaformer}. For AP, we report the IoU thresholds to 0.5, 0.75 and 0.95. For AR, as the small number of forged segments, we set the number of proposals to 100, 50, 30, 20, 10, and 5 with IoU thresholds [0.5:0.05:0.95], respectively.

\subsubsection{Implementation Details}
UniCaCLF is trained end-to-end by Adam optimizer with a learning rate of $1e-3$, a batch size of 8, betas of 0.9 and 0.999, and epsilon of $1e-8$. The number of CaP layer $L$ is $6$. To produce the instant-level label for each scale level in the feature pyramid, we adopted the corresponding scale average pooling operation to downsampling the original instant-level label. If a segment contains more than $40\%$ forged instants, it will be labeled as a forged segment.

For LAV-DF, TVIL, HAD, and Psynd datasets, we use the two-stream TSN \cite{wang2016temporal} network pre-trained on Kinetics dataset and BYOL-A \cite{niizumi2021byol} pre-trained on AudioSet to extract video features and audio features, respectively. The dimension of the extracted video features is 4096, while that of the audio features is 2048. For AV-Deepfake1M dataset, we use the pre-trained ResNet50 and Wave2vec \cite{baevski2020wav2vec} to extract the corresponding visual and audio features, respectively. The dimension of the extracted video features is 4096, while that of the audio features is 1024.
\begin{table}[t]
	\caption{Comparison of training parameters and inference speed.  Best results are in bold and second best underlined.}
	\centering
	\label{tab:cost}
	\resizebox{\columnwidth}{!}{
		\fontsize{6}{8}\selectfont
		\begin{tabular}{c|ccc}
			\hline
			Method & FPS $\uparrow$ & $\#$Params $\downarrow$ & GFLOPs $\downarrow$ \\
			\hline
			ActionFormer \cite{zhang2022actionformer} &31.03&9.70M&{\ul 5.56}   \\
			TriDet \cite{shi2023tridet} &{\ul 31.75}&{\ul 8.08}M&5.93   \\
			UMMAFormer \cite{zhang2023ummaformer}  &11.92&45.00M&20.35  \\
			\hline
			UniCaCLF  &\textbf{35.08}&\textbf{7.32}M&\textbf{4.56}\\
			\hline
		\end{tabular}
	}
\end{table}

\begin{table*}[t]
	\caption{Temporal forgery localization results on the HAD dataset. Best results are in bold and second best underlined.}
	\centering
	\label{tab:had}
	\resizebox{0.9\textwidth}{!}{
		\fontsize{6}{8}\selectfont
		\begin{tabular}{c|cccc|ccccc}
			\hline
			\multirow{2}{*}{Method}  & \multicolumn{4}{c|}{AP@IoU(\%)}                                    & \multicolumn{5}{c}{AR@Proposals(\%)}                                                \\
			& 0.5            & 0.75           & 0.95           & Avg.           & 100            & 50             & 20             & 10             & Avg.           \\ \hline
			ActionFormer \cite{zhang2022actionformer}                              & {\ul 99.94}    & {\ul 99.81}    & 81.29          & 93.68          & 98.12          & 98.07          & 98.05          & 98.03          & 98.07          \\
			TriDet \cite{shi2023tridet}                               & \textbf{99.96} & 99.78          & {\ul 85.51}    & {\ul 95.08}    & {\ul 98.76}    & {\ul 98.75}    & {\ul 98.72}    & {\ul 98.71}    & {\ul 98.74}    \\
			UMMAFormer \cite{zhang2023ummaformer}                              & 99.97          & 99.78          & 76.99          & 92.25          & 97.74          & 97.74          & 97.74          & 97.73          & 97.74          \\ \hline
			UniCaCLF                               & 99.92          & \textbf{99.84} & \textbf{91.52} & \textbf{97.09} & \textbf{99.19} & \textbf{99.18} & \textbf{99.18} & \textbf{99.18} & \textbf{99.18} \\ \hline
		\end{tabular}
	}
\end{table*}
\begin{table*}[t]
	\caption{Comparison results on cross-dataset generalization in terms of AP and AR. Best results are in bold and second best underlined.}
	\centering
	\label{tab:cross}
	\resizebox{0.8\linewidth}{!}{
		\fontsize{7}{9}\selectfont
		\begin{tabular}{c|cccc|ccccc}
			\hline
			\multirow{2}{*}{Method}  & \multicolumn{4}{c|}{AP@IoU(\%)}                                    & \multicolumn{5}{c}{AR@Proposals(\%)}                                                \\
			& 0.5            & 0.75           & 0.95           & Avg.           & 100            & 50             & 20             & 10             & Avg.           \\ \hline
			ActionFormer      & 32.59  & 14.32   & {\ul 1.48}   & 16.13     &57.22 &55.32  &54.68  &52.53 &54.94   \\
			TriDet     & {\ul 43.67}  & {\ul 21.02}    & 0.78   & 21.82   &{\ul 60.13}&{\ul 59.75}&{\ul 58.10}&{\ul57.09 }&{\ul 58.76}   \\
			UMMAFormer       & 21.05  & 6.55    & 0.12  & 9.24 &28.73&28.73&28.35&27.34&28.29          \\ \hline
			UniCaCLF      & \textbf{44.05}  & \textbf{33.38}   & \textbf{3.74}   & \textbf{27.06} &\textbf{67.47}&\textbf{66.33}&\textbf{65.95}&\textbf{64.68}&\textbf{66.11} \\ \hline
		\end{tabular}
	}
\end{table*}

\subsubsection{Baseline Methods}
In order to demonstrate the superiority of our model, we select some advanced TFL works for comparison, including MDS \cite{chugh2020not}, AVFusion \cite{bagchi2021hear}, AVTFD \cite{liu2023audio}, BA-TFD+ \cite{cai2023glitch}, UMMAFormer \cite{zhang2023ummaformer}, and MFMS\cite{zhang2024mfms}.Although DiMoDif\cite{koutlis2024dimodif} and Vigo \cite{perez2024vigo} address related tasks, we exclude them from our comparisons due to the lack of released code, which hinders reproducibility. Moreover, they employ different pre-trained feature extraction models, and such variations have a significant impact on detection performance. To ensure fair evaluation, we consider only reproducible methods and adopt unified pre-trained features across all models. Besides, we also select several advanced TAL methods for comparison, including ActionFormer \cite{zhang2022actionformer} and TriDet \cite{shi2023tridet}. Soft-NMS is adopted as post-processing operation if needed.

\subsection{Performance Comparisons}
To evaluate the universal temporal forgery localization ability of UniCaCLF, we test the proposed framework and all competitors' methods on LAV-DF, AV-Deepfake1M, TVIL, and HAD datasets, respectively. These four datasets stand for multimodal forgery, visual forgery and audio forgery in realistic application scenario, respectively. Since some competitors (e.g., ActionFormer, TriDet, UMMAformer, and MFMS, etc.) require the extraction of pre-trained feature, the same pre-training video and audio encoders as our proposed method are used for them. 

\begin{table*}
	\caption{The contribution of different proposed modules in the UniCaCLF. The experimental results are computed on the TVIL dataset. Best results are in bold and second best underlined.}
	\centering
	\label{tab:component}
	\resizebox{0.9\textwidth}{!}{
		\fontsize{6}{8}\selectfont
		\begin{tabular}{ccc|c|cccc|ccccc}
			\hline
			\multirow{2}{*}{HAO} & \multirow{2}{*}{CaCL} & \multirow{2}{*}{ACU}  & \multicolumn{4}{c|}{AP@IoU(\%)}                                    & \multicolumn{5}{c}{AR@Proposals(\%)}                                                \\
			&                       &                                       & 0.5            & 0.75           & 0.95           & Avg.           & 100            & 50             & 20             & 10             & Avg.           \\ \hline
			&                       &                                         & 79.31          & 76.05          & 61.56          & 72.31          & 91.25          & 89.25          & 87.48          & 84.38          & 88.09          \\
			$\checkmark$                    &                       &                                       & {\ul 86.82}    & {\ul 83.55}    & 69.11          & 79.83          & \textbf{93.15} & {\ul 91.70}    & \textbf{90.19} & {\ul 87.84}    & {\ul 90.72}    \\
			$\checkmark$                    & $\checkmark$                     &                                          & 85.46          & 82.69          & {\ul 73.09}    & {\ul 80.41}    & 92.69          & 91.21          & 89.14          & 87.33          & 90.09          \\
			$\checkmark$                    &                       & $\checkmark$                                       & 85.43          & 82.70          & 71.73          & 79.95          & {\ul 93.12}    & 91.36          & 89.61          & 87.82          & 90.48          \\
			$\checkmark$                    & $\checkmark$                     & $\checkmark$                                        & \textbf{87.07} & \textbf{84.75} & \textbf{74.99} & \textbf{82.27} & 92.93          & \textbf{91.83} & {\ul 90.15}    & \textbf{88.43} & \textbf{90.84} \\ \hline
		\end{tabular}
	}
\end{table*}

\begin{table}[t]
	\caption{Design of the number of pyramid levels.}
	\centering
	\label{tab:levels}
	\resizebox{\columnwidth}{!}{
		\fontsize{8}{10}\selectfont
		\begin{tabular}{c|cc|c|cc}
			\hline
			\multirow{2}{*}{$\#$Levels} & \multicolumn{2}{c|}{AP@IoU(\%)} & \multirow{2}{*}{$\#$Levels} & \multicolumn{2}{c}{AP@IoU(\%)} \\
			& 0.95           & Avg.            &                         & 0.95           & Avg.           \\ \hline
			1                       & 73.29          & 81.74          & 2                       & 72.86          & 79.60          \\
			3                       & 72.75          & 79.97          & 4                       & 73.25          & 81.39         \\
			5                       & 73.33          & 81.03          & 6                       & 74.99          & 82.27         \\
			7                       & 74.68          & 82.48          & 8                       & 74.03          & 81.65              \\ \hline
		\end{tabular}
	}
\end{table}

\subsubsection{Multimodal Modified Scenario}
As shown in Table \ref{tab:fullset}, UniCaCLF can outperform all compared methods under both AP and AR and achieves $16.39\%$ and $2.37\%$ performance gains than state-of-the-art results in terms of AP@0.95 and AR@10. Although MDS was the first to introduce the concept of TFL and attempt to solve TFL problem, it is still essentially a frame-based forgery detection method. Similarly, AVFusion is also a frame-based forgery detection method, and thus neither method can predict boundaries for forged segments. Although ActionFormer and TriDet are both TAL approaches, they perform much better than BA-TFD+ and AVTFD by a major margin under AP@0.95 and AR as they have more advanced detection and localization headers that generate precise proposals for fake segments. In addition, both BA-TFD+ and AVTFD lack the effective feature extraction methods to extract discriminative instant features from raw video and audio, which also leads to their poor performance in LAV-DF dataset. UMMAFormer, a SOTA TFL method that utilizes feature reconstruction method to distinguish forgery traces and detect abnormal instants, achieves better performance than ActionFormer and TriDet, but it only focuses on the feature distribution of real instants and ignores the relative difference between abnormal instants and real instants, still having much room for improvement. In contrast, the proposed UniCaCLF not only uses anomaly detection method to identify forged instants, but also enlarges the relative distance between genuine instants and forged instants via the context aware contrastive learning, resulting in highly discriminative instant features. Therefore, UniCaCLF can have better forgery localization performance. 

Moreover, we also compare the proposed UniCaCLF with previous state-of-the-art approached on the more large-scale AV-Deepfake1M dataset. Compared with LAV-DF, AV-Deepfake1M has a larger scale of forged samples, which facilitates the training of the model for higher localization performance. However, it also incorporates large language models such as ChatGPT to model diverse tampering durations and tampering types, posing a greater challenge to localization. As shown in Table \ref{tab:1m}, UniCaCLF still maintains its superior performance and achieves AP@0.95 of $52.71\%$ and AR@5 of $92.32\%$, outperforming all pervious methods by a large margin ($11.54\%$ at AP@0.95 and $2.6\%$ at AR@5).  This is because benefiting from CaCL, UniCaCLF focuses on modeling anomalous information within the sample, independent of the type of forgery. The strong ability to distinguish between fake and genuine instants makes UniCaCLF better able to handle diverse forgery scenarios.

\begin{figure*}[htbp]
	\begin{center}
		\includegraphics[width=0.8\linewidth]{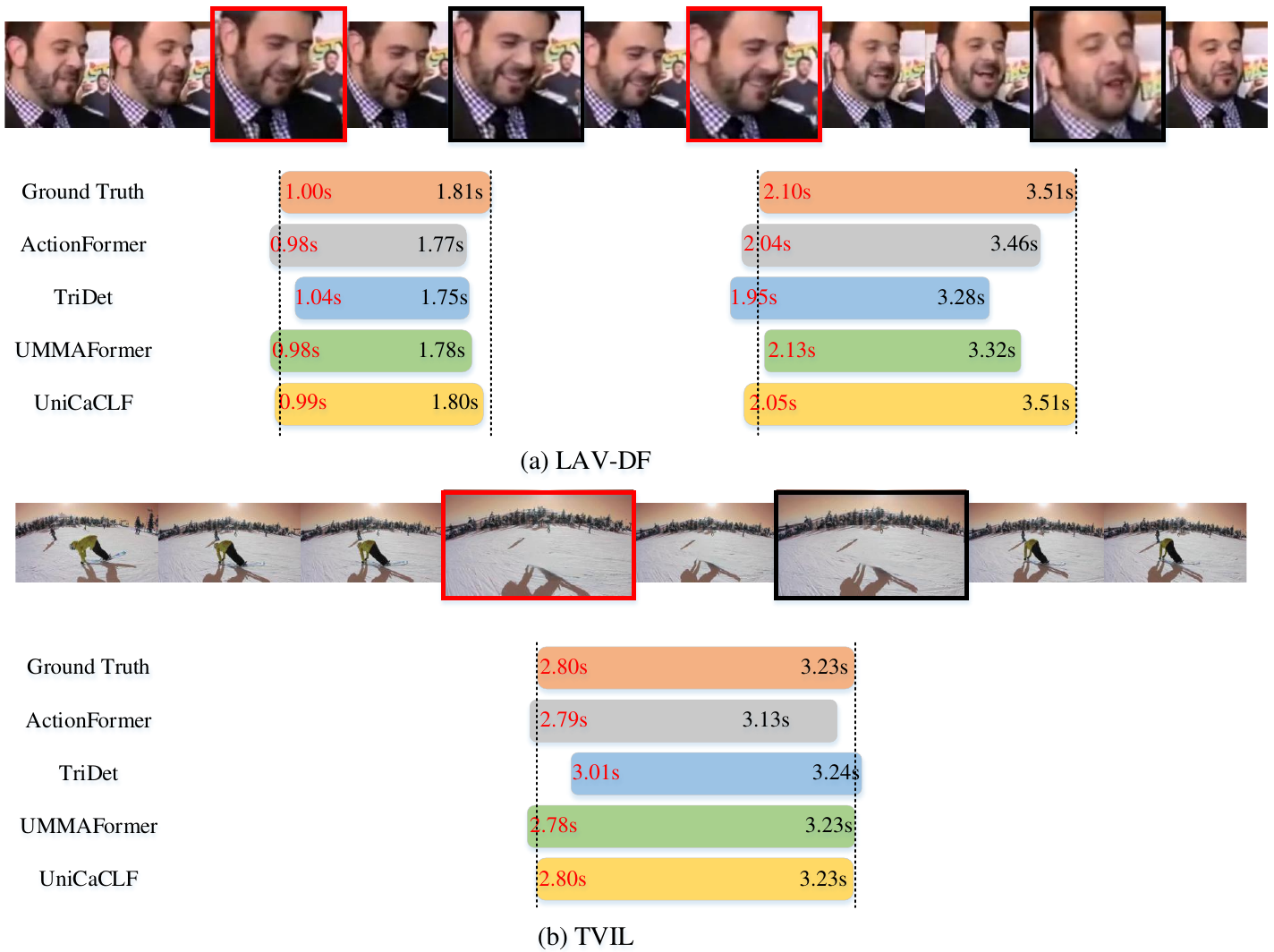}
	\end{center}
	\caption{Visualization results of the LAV-DF and TVIL datasets. The start and end timestamps (in seconds) of the forged segments are highlighted in red and black text, respectively.}
	\label{fig:visual}
\end{figure*}


\subsubsection{Video-only Modified Scenario}
In order to assess the ability of UniCaCLF to cope with generalized forgeries, (\textit{i.e.,} not limited to face forgery), we conduct experiments on the TVIL dataset. As shown 	in Table~\ref{tab:tvil}, relying on the powerful feature representation capabilities, UniCaCLF achieves an average AP of $82.27\%$ and an average AR of $90.84$, outperforming all competitors by at least $2.92\%$ and $1.07$. This boost is significant as the result is averaged across many IoU thresholds, including those tight ones \textit{e.g.}, 0.95. UniCaCLF ranks 2nd with a gap of $2.33\%$ and $1.24\%$ to the UMMAFormer under AP@0.5 and AP@0.75. The possible reason is that CaP layer discards the self-attention mechanism and lacks some ability to capture temporal information. In contrast, ActionFormer and UMMAFormer obtain good forgery localization performance due to the superior temporal information modeling capabilities of the Transformer structure. TriDet lacks both the robust temporal information modeling capability like the Transformer and the ability to effectively mine anomalies in the video, resulting in subpar performance on TVIL dataset.

In addition to model performance, we also compare the parameters and computational cost of all competitors. We start the experiment with 50 warm-up rounds to stabilize GPU performance, followed by 300 iterations for processing the sample of TVIL. As presented in Table~\ref{tab:cost}, our method has the fewest number of trainable parameters ($\#$Params) and floating point operations (GFLOPs). For inference times, our method can process $35.08$ samples per second, outperforming all competitors. The observed fast inference speed stems from that the proposed CaP layer eschews intensive matrix operations and employs the heterogeneous activation operation to measure the similarity of the instant features. These results demonstrate that our approach achieves a favorable trade-off between performance and computational cost.

\subsubsection{Audio-only Modified Scenario}
For the audio-only modified dataset HAD, the comparison results presents in Table~\ref{tab:had}. UniCaCLF shows promising results, outperforming the strong baseline TriDet by 6.01$\%$ under AP@0.95 and 0.47$\%$ under AR@10. Compared to the rich semantic information contained in video, audio is simpler and purer. In other words, audio features have better task-specific performance compared to visual features. Consequently, all models exhibit good detection performance on the HAD dataset.

\subsubsection{Cross-dataset Comparisons}
In this section, we train the models on the HAD dataset and test on the Psynd dataset to evaluate the generalization of models. Comparisons under AP and AR metrics are shown in Table~\ref{tab:cross}. UniCaCLF achieves an average AP of $27.06\%$ and an average AR of $66.11\%$, exceeding the competitors by a large margin $5.24\%$ and $7.35\%$. This is reasonable as UniCaCLF is trained in a supervised sample-by-sample manner. CaCL makes UniCaCLf more focused on modeling the relative differences between real and fake instant features within a sample, immune to the adverse effects of different tampering operations, which allows the UniCaCLF to extract more general anomaly detection features.

\subsection{Ablation Study}
\subsubsection{Main Component Analysis}
We conduct comprehensive ablation studies on TVIL dataset to further explore the effectiveness of the proposed framework components, \textit{i.e.}, HAO, ACU, and CaCL, as listed in Table~\ref{tab:component}. When UniCaCLF does not have these three components, we refer to as the baseline. In other words, baseline is the CaP layer of UniCaCLF replaced with a convolution-based residual block and is no supervised by CaCL. From Table~\ref{tab:component}, we can find that only introducing HAO boosts the AP@0.95 by 7.55$\%$. This suggests that HAO can effectively increase the feature discriminability between forged instants and genuine instants. With the addition of CaCL or ACU, the framework performance can be further improved. In particular, CaCL results in a major performance improvement by $3.98\%$ at AP@0.95, indicating that the proposed supervised sample-by-sample contrastive learning method is critical for the UniCaCLF. Reducing the variations of genuine instant feature in the embedding space can be effective in increasing the difference between genuine and forged instants. ACU also does provide a more positive global context that ensure the precision of the context-aware contrastive objective construction, thus improving the performance of UniCaCLF. These three proposed components are complementary and can work together to maximize the performance of the UniCaCLF.

\subsubsection{Study on the number of pyramid levels}
As shown in Fig.~\ref{fig:cos}, the cosine similarity of Cap layer increases greatly when layers grow, and the SGP layer performs better when the number of layers is greater than four. It seems that the fewer the Cap layers, the better the model localization performance. In order to set an appropriate number of pyramid layers, we start the ablation from the feature pyramid with $8$ different levels on TVIL dataset. As shown in Table~\ref{tab:levels}, we can find that the proposed framework is insensitive to the number of pyramid layers, and competitive localization results can be obtained with only one CaP layer. With more levels (\textit{i.e.}, levels more than $6$), the proposed framework can also achieve better performance. That is because the Cap layer itself lacks temporal modeling capability, and a multi-levels feature pyramid can provide some temporal information to compensate for this shortcoming.


\subsection{Localization Visualization Analysis}
As shown in Fig.~\ref{fig:visual}, we present the localization results of the UniCaCLF, ActionFormer, TriDet and UMMAFormer on the LAV-DF and TVIL datasets, respectively. Obviously, our method can predict more accurate the start and end instants than all competitors on LAV-DF and TVIL datasets, indicating that the instant features extracted by UniCaCLF are more discriminative than those extracted by all competitors.

\section{Conclusions}\label{sec:con}
In this paper, we propose a novel universal context-aware contrastive learning framework named UniCaCLF for TFL task. The UniCaCLF consists of three core components: the heterogeneous activation operation (HAO), the adaptive context updater (ACU), and the context-aware contrastive loss (CaCL). HAO and ACU cooperate with each other to enhance the feature representation of forged instants. CaCL and ACU cooperate with each other to narrow the distance between the genuine instant and the global context, and widen the distance between the forged instant and the global context, thus further boosting inter-class distinguishability in the embedding space. Eventually, UniCaCLF can learn more discriminative instant features for TFL task. Extensive experimental results demonstrate the superiority of UniCaCLF in the forensics for different modal forgery scenarios. 
%

\bibliographystyle{IEEEtran}
\bibliography{egbib}

\end{document}